\documentclass{article}

\usepackage[preprint]{cs224r_2025}
\usepackage[utf8]{inputenc} 
\usepackage[T1]{fontenc}    
\usepackage{hyperref}       
\usepackage{url}            
\usepackage{booktabs}       
\usepackage{amsfonts}       
\usepackage{nicefrac}       
\usepackage{microtype}      
\usepackage{xcolor}         
\usepackage{amsmath}
\usepackage{graphicx}       
\usepackage{lipsum}         
\usepackage{multirow}
\usepackage{longtable}

\title{MEMBOT: Memory-Based Robot in Intermittent POMDP}

\author{%
  Eyan Noronha\thanks{Equal contribution.} \\
  Department of Computer Science \\
  Stanford University \\
  \texttt{eyannoro@stanford.edu}
  \And
  Yousef (Youzhi) Liang\footnotemark[1] \\
  Department of Computer Science \\
  Stanford University \\
  \texttt{youzhil@stanford.edu}
}

\begin{document}

\maketitle

\begin{abstract}
Robotic systems deployed in real-world environments often operate under conditions of partial and often intermittent observability, where sensor inputs may be noisy, occluded, or entirely unavailable due to failures or environmental constraints. Traditional reinforcement learning (RL) approaches that assume full state observability are ill-equipped for such challenges. In this work, we introduce MEMBOT, a modular memory-based architecture designed to address intermittent partial observability in robotic control tasks. MEMBOT decouples belief inference from policy learning through a two-phase training process: an offline multi-task learning pretraining stage that learns a robust task-agnostic latent belief encoder using a reconstruction losses, followed by fine-tuning of task-specific policies using behavior cloning. The belief encoder, implemented as a state-space model (SSM) and a LSTM, integrates temporal sequences of observations and actions to infer latent state representations that persist even when observations are dropped. We train and evaluate MEMBOT on 10 robotic manipulation benchmark tasks from MetaWorld and Robomimic under varying rates of observation dropout. Results show that MEMBOT consistently outperforms both memoryless and naively recurrent baselines, maintaining up to 80\% of peak performance under 50\% observation availability. These findings highlight the effectiveness of explicit belief modeling in achieving robust, transferable, and data-efficient policies for real-world partially observable robotic systems.
\end{abstract}

\section{Introduction}

Autonomous decision-making in robotics often confronts a fundamental challenge: the world is only partially observable. Real-world robots must operate with sensors that are frequently noisy, incomplete, or unreliable—due to occlusion, limited fields of view, bandwidth constraints, or hardware failures. This sensory degradation leads to uncertainty about the true state of the environment, making it difficult to plan effectively or act reliably. While many control and reinforcement learning (RL) methods assume full observability for tractability, this assumption is rarely valid outside of controlled laboratory conditions.

To address such uncertainty, the Partially Observable Markov Decision Process (POMDP) framework offers a principled approach. In a POMDP, the agent does not directly observe the true environmental state but must instead infer a belief over possible states based on its observation and action history. Solving a POMDP thus requires maintaining internal memory and reasoning over that memory to make optimal decisions. However, classical POMDP solvers scale poorly in high-dimensional and continuous domains, especially in robotics, where observations and action spaces are complex and dynamic.

Modern deep RL has increasingly turned to differentiable sequence models—such as Long Short-Term Memory (LSTM) networks, Simple State-Space Models (SSM) and transformer-based attention mechanisms—to approximate this belief state. These models learn to encode temporal context from raw observation sequences, enabling agents to act based on a learned latent history. While this approach has improved robustness under partial observability, most existing work assumes consistent access to sensor streams. In contrast, many real-world deployments suffer from intermittent sensor failures—where entire observation modalities are lost for periods of time due to occlusion, network latency, or malfunction. Handling such dropouts requires more than generic recurrence; it demands belief representations that persist and remain informative even in the absence of immediate sensory input.

This project introduces MEMBOT: a memory-based architecture designed specifically for robotic control under intermittent partial observability. MEMBOT decouples belief state estimation from policy learning, using a shared SSM-based observer and a LSTM-based observer to maintain temporal belief representations across tasks. The belief model is pretrained offline using expert demonstrations with both behavior cloning and reconstruction losses, ensuring robustness to missing observations. Task-specific policies are then fine-tuned online via actor-critic learning, leveraging the pretrained belief encoder to enable fast adaptation and robust action selection.

Through systematic experiments on robotic manipulation benchmarks—including MetaWorld and Robomimic—under varying observation dropout conditions, we show that MEMBOT substantially outperforms memoryless and naively recurrent baselines. The architecture maintains 65–80\% of peak performance even when 50\% of observations are randomly dropped—compared to baseline models which degrade to 10–30\% under the same conditions.

Overall, this work contributes a modular framework for robust policy learning in real-world POMDPs with intermittent sensing. By explicitly modeling belief with memory and decoupling it from policy logic, MEMBOT provides a practical step toward resilient, deployable autonomous systems capable of functioning under real-world sensory constraints.

\section{Related Work}

This section briefly surveys the research landscape on reinforcement learning in robotics involving partial observability, sequence models for policy memory, and approaches to sensor-level dropout. The reviewed works span from foundational studies on POMDPs to modern applications of LSTM and transformer-based policies, including efforts to increase robustness to sensor failures in real robotic environments.

\subsection{Partially Observable Markov Decision Processes}

Many real-world robotic applications naturally exhibit partial observability due to limited, noisy, or ambiguous sensor data. This motivates the formulation of these tasks as Partially Observable Markov Decision Processes (POMDPs), where the agent must infer hidden states from a sequence of observations and actions. Several studies \citep{kastner2021autonomous, zhang2015policy, hausknecht2015deep, fang2019scene} explicitly recognize and address the challenges of POMDPs in the context of robotic navigation and control. These works demonstrate that traditional, memoryless policies are inadequate in such settings, and that specialized approaches are necessary to maintain and utilize historical information for optimal decision-making. POMPDPs are intractable to optimally solve by inferring a belief distribution over the underlying state for all but the smallest of problems~\citep{basich2022planning, zhong2023optimizing}. Offline approximation methods have been developed to scale better than exact methods to larger problems. However, designing efficient and robust online algorithms remains challenging,  particularly under intermittent POMDPs~\citep{avrachenkov2025constrained}.

\subsection{Sequence Models}

To address the challenges of memory in POMDPs, a variety of sequence modeling architectures have been integrated into reinforcement learning pipelines. Recurrent neural networks, particularly Long Short-Term Memory (LSTM) networks, have been widely adopted to summarize past observations when learning policies in partially observed environments \citep{aikins2024robust, parsbin2023visual, jiang2022multi, li2024deep}. These models enable the policy to condition actions on internal hidden states representing observation histories, resulting in consistent improvements over feedforward baselines. Additionally, transformer-based memory architectures—such as the Scene Memory Transformer \cite{fang2019scene}—have begun to demonstrate potential for handling longer-range temporal dependencies in complex navigation tasks.
Alternative approaches include explicit continuous-valued memory states, as seen in guided policy search frameworks \citep{zhang2015policy}, and adaptations for dynamic environments using recurrent neural networks \cite{cho2023low}. Across these studies, sequence models play a central role in facilitating robust policy learning where memory of prior events is essential.

\subsection{Sensor-Level Dropout}

Robustness to sensor failures, such as intermittent sensor dropout or occlusion, is a growing concern in real-world robotic deployment. Some recent works have addressed this by explicitly simulating sensor "flicker," blackout, or providing missing/noisy observations during learning and evaluation \citep{aikins2024robust, hausknecht2015deep}. For example, LSTM-augmented reinforcement learning policies have been shown to outperform both feedforward and classical filtering controllers under such conditions \citep{aikins2024robust}. However, across the surveyed literature, only a limited number of studies have systematically targeted sensor-level dropout as a primary experimental focus. Most works rely on the inherent temporal integration capabilities of sequence models to confer partial robustness to real-world noise or incomplete sensing \citep{han2023multimodal, jiang2022multi}, but comprehensive benchmarking under explicit sensor occlusion scenarios remains relatively sparse.

Despite significant progress, existing methods fall short in systematically addressing intermittent observability—a common and critical failure mode in real-world robotics. Most prior works either assume consistent, albeit noisy, observations or evaluate robustness only under synthetic noise without modeling temporally correlated sensor outages. Furthermore, many sequence models tightly couple memory and policy, limiting modularity and transferability across tasks. In contrast, our approach introduces a decoupled belief-policy architecture (MEMBOT) that explicitly handles observation dropout through dedicated latent belief modeling. By jointly training a shared memory-based encoder offline and fine-tuning task-specific policies online, MEMBOT achieves robust inference and action selection under intermittent sensing while maintaining transferability and sample efficiency. This modular design bridges key gaps in the literature, offering a scalable solution for POMDPs with practical sensor constraints.

\section{Method}
\subsection{Problem Formulation}

\begin{figure}[htbp]
\centering
\begin{minipage}{0.53\textwidth}
    \centering
    \includegraphics[width=\linewidth]{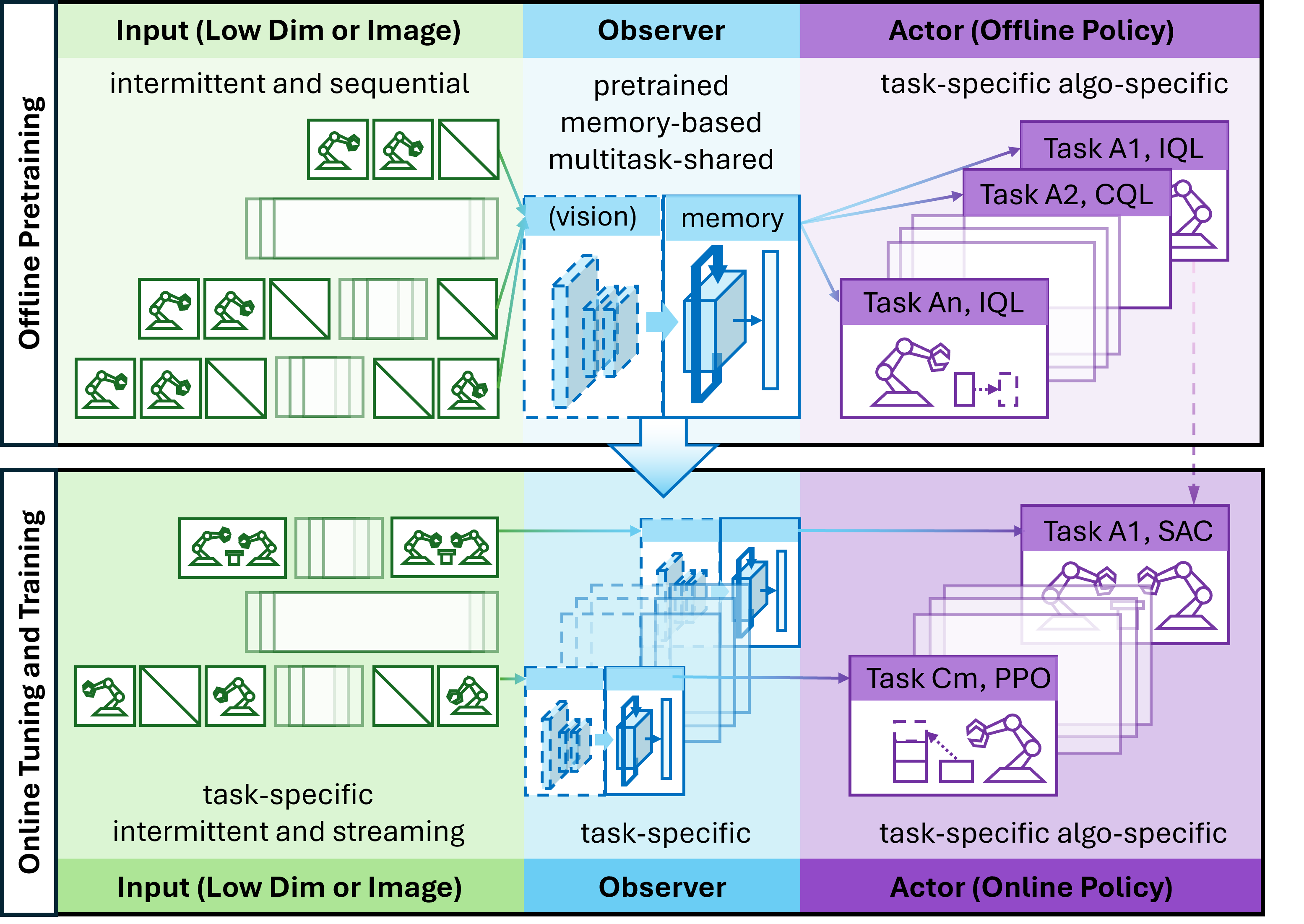}
    \caption{System architecture of MEMBOT, a two-phase offline-online training methodology with a shared latent belief encoder and task-specific policy heads.}
    \label{fig:membot_architecture}
\end{minipage}
\hfill
\begin{minipage}{0.42\textwidth}
    \centering
    \small  
    \begin{tabular}{cl}
    \toprule
    \textbf{Notation} & \textbf{Description} \\
    \midrule
    \multicolumn{2}{l}{\textit{Problem Formulation}} \\
    $\mathcal{T}_i$ & Task $i$ as POMDP \\
    $\mathcal{S}, \mathcal{A}, \mathcal{O}$ & State, action, observation spaces \\
    $T_i, R_i, \Omega_i$ & Transition, reward, observation \\
    $\gamma$ & Discount factor \\
    $m_t$ & Observation availability indicator \\
    $p$ & Observation probability \\
    \midrule
    \multicolumn{2}{l}{\textit{MEMBOT Architecture}} \\
    $f_\phi$ & Observation encoder\\
    $g_\psi$ & Memory-based observer\\
    $\pi_\theta$ & Task-specific policy \\
    $F_{\phi,\psi}$ & Complete belief encoder \\
    $e_t$ & Encoded observation \\
    $b_t$ & Belief state \\
    $h_t$ & SSM/LSTM hidden state \\
    \bottomrule
    \end{tabular}
    \vspace{0.5em}
    {\footnotesize Table 1: Key notation for MEMBOT.}
    \label{tab:notation}
\end{minipage}
\end{figure}

We model robotic object manipulation tasks under intermittent partial observability as a collection of Partially Observable Markov Decision Processes (POMDPs). Table~\ref{tab:notation} shows the key notations for this problem formulation and MEMBOT architecture. Each task $\mathcal{T}_i$ in our training set $\mathcal{T}_{\text{train}}$ is defined as:

\begin{equation}
\mathcal{T}_i = (\mathcal{S}, \mathcal{A}, \mathcal{O}, T_i, R_i, \Omega_i, \gamma).
\end{equation}


The key challenge we attempt to address is the intermittent observability, where observations $o_t$ may be unavailable at certain timesteps due to hardware malfunctions, environment realities, physical obstruction or system optimization trade-offs, as discussed in previous sections. We model this scenario by introducing observation availability indicators $m_t \in \{0,1\}$, where $m_t = 0$ indicates missing observations. The probability of $m_t = 1$ is the observation probability $p$. The actual observation received by the agent is:

\begin{equation}
\tilde{o}_t = m_t \cdot o_t + (1 - m_t) \cdot \mathbf{0}
\end{equation}

where $\mathbf{0}$ represents a zero vector of appropriate dimensionality, subject to the specific task and environment. The observations can be low-dimensional states of robots with the states of objects or vision observations such as images from various perspectives.

\subsection{MEMBOT Architecture}

Our proposed MEMBOT architecture consists of three key modules designed to address intermittent observability while enabling transfer learning across different tasks within the scope of robotic object manipulation:

\textbf{Observation Encoder ($f_\phi$):} A neural network that maps raw observations to latent representations:
\begin{equation}
e_t = f_\phi(\tilde{o}_t)
\end{equation}

This component abstracts raw sensory inputs into a consistent latent space, which is essential for handling diverse observation modalities and reducing the dimensionality of subsequent temporal processing. The architecture of this component depends on the modality of the input. For low-dimensional input, this component is a multi-layer perceptron (MLP); for image input, this component is a MLP followed by a Convolutional Neural Network (CNN). In our current implementation, we use a MLP with ReLU activations and layer normalization for the testings, which will be detailed in the Experiment Setup section.

\textbf{Memory-based Observer ($g_\psi$):} We design a sequence model that maintains the latent temporal belief states by integrating current encoded observations with historical information:
\begin{equation}
b_t, h_t = g_\psi(e_t, h_{t-1})
\end{equation}


This temporal integration is crucial for maintaining coherent latent beliefs about the environment and object states when observations are intermittently unavailable, allowing the agent to make informed decisions based on historical context. We implement this with two memory-based models. One is using Long Short-Term Memory (LSTM) networks and its variations with varied length of historical observations.

To explore more lightweight alternatives, we also consider a simple linear state-space model (SSM), where the belief state is updated via a fixed learned transformation~\citep{becker2024kalmamba}. Specifically, we use a linear state update mechanism with a non-linear activation to propagate temporal information:
\begin{equation}
    h_t = \tanh(W_h h_{t-1} + W_o o_t + b)
\end{equation}
where $h_t$ is the hidden state at time $t$, $o_t$ is the observation, and $W_h$, $W_o$, and $b$ are learnable parameters. The updated state $h_t$ is then projected to a belief vector which is used by the policy and value networks.

\textbf{Task-specific Policy ($\pi_\theta$):} We design a neural network that maps the latent belief states to action distributions:
\begin{equation}
a_t \sim \pi_\theta(\cdot \mid b_t)
\end{equation}

The architecture of this component is a regular MLP design. Operating on belief states rather than raw observations enables the policy to make decisions based on a comprehensive understanding of the current situation, incorporating both immediate and historical information. Our implementation uses a MLP with tanh output activation to produce continuous action distributions.

As a summary of the three components discussed above, the complete belief encoder is defined as:
\begin{equation}
F_{\phi,\psi}(h_{t-1}, \tilde{o}_t) = g_\psi(f_\phi(\tilde{o}_t), h_{t-1})
\end{equation}

\subsection{Two-Phase Training Methodology}

We use a two-phase training approach that separates belief state learning from policy optimization, enabling robust performance under varying observation conditions and efficient transfer across tasks for MEMBOT training.

\subsubsection{Phase 1: Offline Belief Encoder Pretraining}

In the first phase, we pretrain a shared latent belief encoder using expert demonstration data collected across multiple tasks. This phase combines behavior cloning with reconstruction objectives to learn robust latent belief representations.

The behavior cloning loss encourages the policy to imitate expert actions:
\begin{equation}
\mathcal{L}_{\text{BC}} = -\mathbb{E}_{(b_t, a_t^*) \sim \mathcal{D}_{\text{expert}}} \left[ \log \pi_\theta(a_t^* \mid b_t) \right],
\end{equation}

where $a_t^*$ represents expert actions and $\mathcal{D}_{\text{expert}}$ is the expert demonstration dataset. To ensure that belief states capture essential environmental information, we introduce a reconstruction loss that trains a decoder network $p_\xi(o_t \mid b_t)$ to reconstruct observations from belief states:
\begin{equation}
\mathcal{L}_{\text{recon}} = -\mathbb{E}_{(b_t, o_t) \sim \mathcal{D}_{\text{expert}}} \left[ \log p_\xi(o_t \mid b_t) \right].
\end{equation}

The joint pretraining objective is:
\begin{equation}
\mathcal{L}_{\text{pretrain}} = \mathcal{L}_{\text{BC}} + \lambda \cdot \mathcal{L}_{\text{recon}},
\end{equation}

where $\lambda$ controls the relative importance of reconstruction. This joint objective ensures that the belief encoder learns representations that are both useful for action prediction and capture sufficient information about the underlying environment state. During pretraining, we can simulate intermittent observability by randomly masking observations in the demonstration data with probability $p_{\text{mask}}$.

\subsubsection{Phase 2: Online Task-specific Fine-tuning}

In the second phase, we fine-tune the pretrained system on specific tasks using Soft Actor-Critic (SAC) \citep{haarnoja2018soft}. Our innovation here is that we allow joint optimization of both the policy and the belief encoder, enabling task-specific adaptation while preserving the robust temporal reasoning capabilities learned during pretraining. We have critics updates, actor updates and joint encoder-policy updates. 

\textbf{Critic Updates:} We use an ensemble of critic networks $\{Q_{\psi_k}\}_{k=1}^N$ to improve stability and reduce overestimation bias. The critic targets are computed as:

\begin{equation}
\hat{Q}_t = r_t + \gamma \min_{k \in \{1,\ldots,N\}} Q_{\psi_k}(b_{t+1}, a_{t+1})
\end{equation}

where $a_{t+1} \sim \pi_\theta(\cdot \mid b_{t+1})$ is sampled from the current policy. The critic loss is:

\begin{equation}
\mathcal{L}_{\text{critic}} = \frac{1}{N} \sum_{k=1}^N \mathbb{E}_{(b_t, a_t, r_t, b_{t+1}) \sim \mathcal{B}} \left[ \left( Q_{\psi_k}(b_t, a_t) - \hat{Q}_t \right)^2 \right]
\end{equation}

where $\mathcal{B}$ represents the replay buffer containing belief state transitions.

\textbf{Actor Updates:} The policy is updated to maximize expected returns while maintaining exploration through entropy regularization:

\begin{equation}
\mathcal{L}_{\text{actor}} = -\mathbb{E}_{b_t \sim \mathcal{B}, a_t \sim \pi_\theta(\cdot \mid b_t)} \left[ \frac{1}{N} \sum_{k=1}^N Q_{\psi_k}(b_t, a_t) - \log \pi_\theta(a_t \mid b_t) \right].
\end{equation}


\textbf{Joint Encoder-Policy Updates:} We allow gradients to flow through the latent belief encoder during actor updates, enabling joint optimization of the latent belief observer including the MLP component and memory-based component. This joint optimization allows the belief encoder to adapt to task-specific requirements while maintaining its core temporal reasoning capabilities.

\begin{equation}
\nabla_{\phi,\psi,\theta} \mathcal{L}_{\text{actor}} = \nabla_{\phi,\psi,\theta} \mathbb{E}_{b_t, a_t} \left[ -\frac{1}{N} \sum_{k=1}^N Q_{\psi_k}(f_{\phi,\psi}(h_{t-1}, \tilde{o}_t), a_t) \right].
\end{equation}

The modular design of MEMBOT enables effective transfer learning across robotic tasks. The shared belief encoder $f_{\phi,\psi}$ learns task-agnostic representations of temporal dynamics and observational patterns, while task-specific policies $\pi_\theta^{(i)}$ adapt to particular reward structures and action requirements. For deployment on a new task $\mathcal{T}_{\text{new}}$, we initialize the belief encoder with pretrained parameters and fine-tune only the policy network, significantly reducing the sample complexity compared to training from scratch. The reconstruction loss during pretraining ensures that the learned belief states capture sufficient environmental information to support diverse downstream tasks.

\subsection{Implementation Details}

We efficiently process sequences of varying lengths during training, we employ packed sequence processing in our memory-based component implementation. For a batch of sequences $\{s^{(i)}\}_{i=1}^B$ with lengths $\{\ell^{(i)}\}_{i=1}^B$, we use PyTorch's packed sequence utilities. This approach ensures computational efficiency by avoiding unnecessary computations on padded elements while maintaining proper gradient flow through the temporal dependencies.

\begin{equation}
\text{packed\_input} = \text{pack\_padded\_sequence}(\{s^{(i)}\}, \{\ell^{(i)}\}, \text{batch\_first=True}). 
\end{equation}

Our implementation uses a three-layer MLP for observation encoding with ReLU activations and layer normalization. The LSTM component has a hidden dimension of 128 with a single layer to balance representational capacity with computational efficiency. The State-Space Model (SSM) models the hidden state of dimension 128 and uses a recurrent linear architecture with a $\tanh$ activation function to reconstruct the observations. The policy and critic networks are implemented as three-layer MLPs with hidden dimensions of 256.

During online training, we maintain a replay buffer of belief state transitions and use prioritized experience replay to improve sample efficiency. The reconstruction weight $\lambda$ is set to 1.0 during pretraining and reduced to 0.1 during fine-tuning to balance belief state quality with task-specific optimization.

To simulate realistic intermittent observability patterns, we vary the observation masking probability $p_{\text{mask}}$ during training, exposing the system to different levels of observation availability. This curriculum approach helps develop robust belief maintenance capabilities across diverse deployment scenarios.

\section{Experimental Setup}
\subsection{Evaluation Environments}
We use two sets of benchmark suites, including MetaWorld benchmark suite and Robomimit benchmark to evaluate MEMBOT on a diverse set of robotic manipulation tasks.

MetaWorld benchmark suite provides standardized continuous control environments for multi-task and meta-learning research\citep{yu2020meta}. Our experimental evaluation focuses on five representative tasks that span different manipulation primitives and difficulty levels: Drawer-Close-v2 (closing an open drawer by grasping and pushing the handle), Handle-Press-v2 (pressing down on a handle mechanism), Plate-Slide-v2 (pushing a plate to a target location on a table surface), Push-v2 (pushing a puck to a designated goal position), and Reach-v2 (moving the end-effector to a target location). Each task uses a 7-DOF robotic arm with continuous action spaces representing joint velocities. The observation space consists of 39-dimensional vectors containing proprioceptive information (joint positions, velocities) and task-relevant object states (positions, orientations). All tasks use sparse binary rewards, where the agent receives a reward of 1.0 upon successful task completion and 0.0 otherwise.

Also, we conduct experiments on five tasks from the Robomimic benchmark suite \citep{mandlekar2021matters}, which provides high-quality demonstration datasets and standardized evaluation protocols for imitation learning research. These tasks include Lift (picking up a cube and lifting it above a target height), Can (picking up a can and placing it in a target bin), Square (inserting a square peg into a square hole), Transport (moving objects between containers), and Tool Hang (hanging a tool on a rack). The Robomimic tasks feature more complex visual observations and longer horizons compared to MetaWorld, providing additional validation of MEMBOT's scalability to challenging manipulation scenarios with richer sensory inputs.





\subsection{Training Details}

Our training consists of two phases. During offline pretrained, we pretrain the shared belief encoder using expert demonstration data collected from each task. Expert trajectories are generated using the provided MetaWorld scripted policies, which achieve near-optimal performance on each task. For each environment, we collect 60 demonstration episodes of length 40 timesteps, resulting in approximately 2,400 state-action pairs per task. During pretraining, we train for 10,000 iterations using the Adam optimizer with a learning rate of $10^{-4}$. We use a batch size of 256. 

During online fine-tuning, we fine-tune the complete system using SAC on each task individually. We train for 100,000 environment steps with the following hyperparameters: learning rate $10^{-4}$, batch size 256, replay buffer size 100,000, and discount factor $\gamma = 0.99$. The entropy regularization parameter $\alpha$ is automatically tuned using the dual formulation \citep{haarnoja2018soft}. During online training, we reduce the reconstruction weight to $\lambda = 0.1$ to balance belief state quality with task-specific optimization. We use an Update-to-Data (UTD) ratio of 1.0 and perform behavior cloning updates every 2 gradient steps to maintain alignment with expert demonstrations. We understand that we have a limited number of training steps because there are 10 tasks that we need to use for multi-tasking learning and use held-out tasks for testing, and we need to vary the architectures and hyperparameters for evaluations.

\section{Results}

We evaluate MEMBOT, its variations and memoryless baselines using the following three metrics. First, we use success rate. The task average success rate is computed as the fraction of episodes where the agent achieves the task objective as defined in the simulation environment within the episode time limit (50 timesteps). This binary metric directly measures task completion performance. Second, we use the average accumulated rewards. We also report the mean cumulative reward per episode, which provides a more nuanced view of performance quality, especially for partially successful episodes. Third, we define a relative degradation specifically for the evaluation of performance in intermittent POMDP, serving as a crucial metric. To quantify robustness to observation dropout, we compute the relative performance degradation:

\begin{equation}
\text{Relative Degradation} = \frac{\text{Performance}(p_{\text{obs}}) - \text{Performance}(p_{\text{obs}} = 1.0)}{\text{Performance}(p_{\text{obs}} = 1.0)},
\end{equation}

where performance can be either success rate or episode return. This metric quantifies the degradation of performance relative to what is achieved at no dropout observations.

\begin{figure}[h!]
    \centering
    \includegraphics[width=1\linewidth]{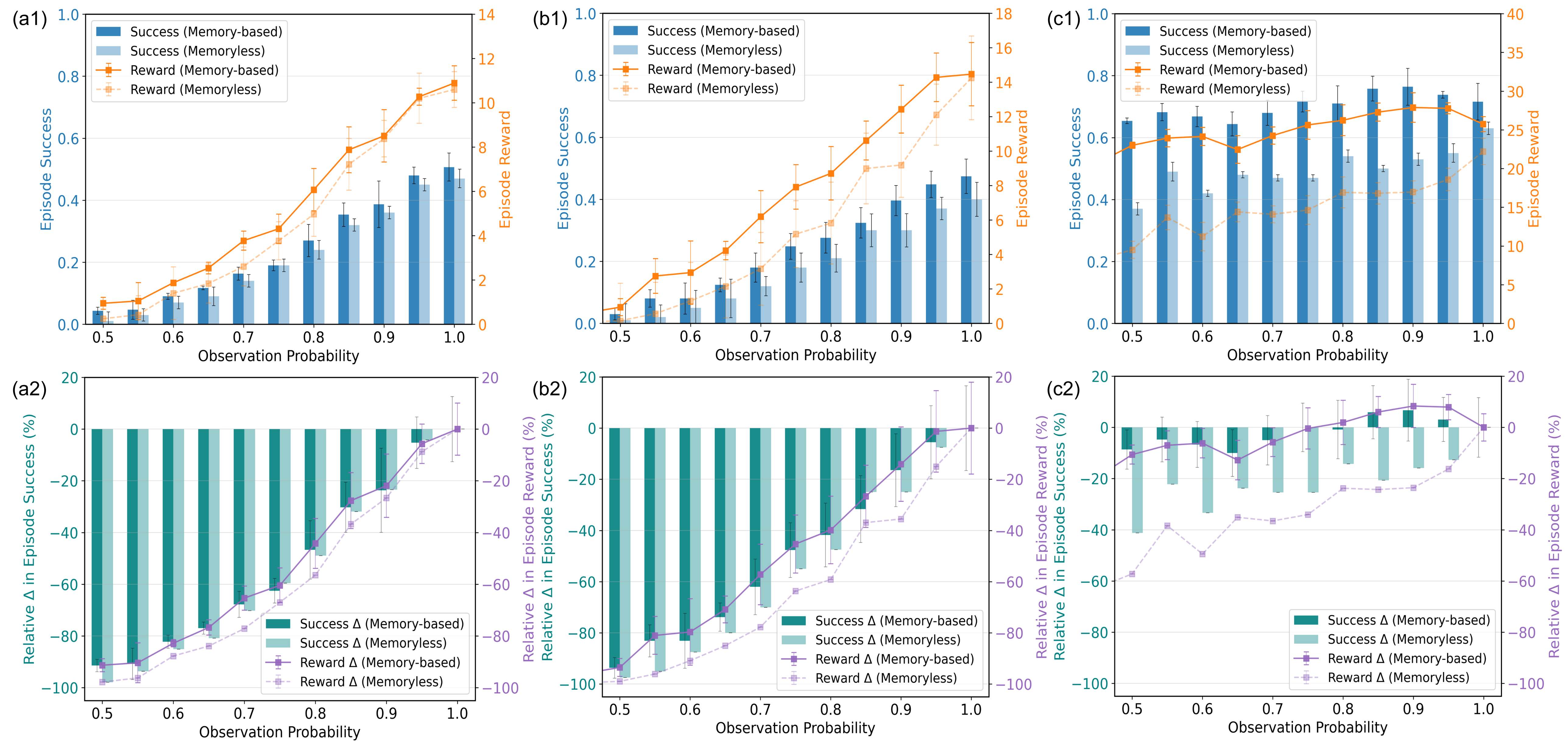}
    \caption{Performance comparison between MEMBOT and memoryless baseline across varying observation probabilities. Top row shows absolute episode success rates (blue bars, left y-axis) and episode rewards (orange lines, right y-axis). Bottom row shows relative performance degradation compared to full observability (observation probability = 1.0), with negative values indicating performance loss. Results shown for three MetaWorld tasks: (a) Plate-Slide-v2, (b) Handle-Press-v2, and (c) Drawer-Close-v2. Observation probability ranges from 0.5 to 1.0 in increments of 0.05. Error bars represent the standard deviation across 3 random seeds with 100 evaluation episodes each. }
    \label{fig:perfvsobs}
\end{figure}

\begin{table}[htbp]
\centering
\caption{Comprehensive ablation study showing progressive improvements through MEMBOT components and comparison with memoryless baselines. Best results are colored in blue.}
\label{tab:comprehensive_ablation}
\resizebox{\textwidth}{!}{%
\begin{tabular}{llc|ccccccccccc}
\toprule
\multirow{2}{*}{Task} & \multirow{2}{*}{Method} & \multirow{2}{*}{Metric} & \multicolumn{11}{c}{Observation Probability} \\
\cmidrule(lr){4-14}
& & & 0.5 & 0.55 & 0.60 & 0.65 & 0.70 & 0.75 & 0.80 & 0.85 & 0.90 & 0.95 & 1.0 \\
\midrule
\multirow{12}{*}{\begin{tabular}[c]{@{}l@{}}Plate-\\Slide\end{tabular}} 
& \multirow{2}{*}{\begin{tabular}[c]{@{}l@{}}MEMBOT\\- State Space\end{tabular}} & Success (\%) & \textcolor{blue}{9.0} & \textcolor{blue}{7.0} & \textcolor{blue}{9.0} & \textcolor{blue}{21.0} & \textcolor{blue}{24.0} & \textcolor{blue}{41.0} & \textcolor{blue}{45.0} & \textcolor{blue}{70.0} & \textcolor{blue}{76.0} & \textcolor{blue}{84.0} & \textcolor{blue}{95.0} \\
& & Reward & \textcolor{blue}{1.38} & \textcolor{blue}{1.06} & \textcolor{blue}{1.95} & \textcolor{blue}{3.51} & \textcolor{blue}{4.89} & \textcolor{blue}{7.66} & \textcolor{blue}{9.54} & \textcolor{blue}{15.33} & \textcolor{blue}{17.30} & \textcolor{blue}{19.80} & \textcolor{blue}{24.33} \\
\cmidrule(lr){2-14}
& \multirow{2}{*}{\begin{tabular}[c]{@{}l@{}}MEMBOT\\- LSTM (Full)\end{tabular}} & Success (\%) & 4.6 & 4.3 & 9.2 & 11.9 & 16.7 & 18.4 & 27.3 & 35.8 & 39.1 & 48.3 & 51.2 \\
& & Reward & 1.0 & 1.0 & 2.0 & 2.6 & 3.9 & 4.1 & 6.3 & 8.1 & 8.7 & 10.5 & 11.1 \\
\cmidrule(lr){2-14}
& \multirow{2}{*}{\begin{tabular}[c]{@{}l@{}}MEMBOT\\- LSTM (Partial-10)\end{tabular}} & Success (\%) & 4.7 & 4.2 & 9.1 & 11.8 & 16.6 & 19.3 & 27.2 & 35.7 & 39.2 & 48.1 & 51.0 \\
& & Reward & 1.0 & 1.0 & 1.9 & 2.5 & 3.8 & 4.4 & 6.2 & 8.0 & 8.8 & 10.4 & 11.0 \\
\cmidrule(lr){2-14}
& \multirow{2}{*}{\begin{tabular}[c]{@{}l@{}}MEMBOT\\- LSTM (No Recon)\end{tabular}} & Success (\%) & 4.2 & 4.8 & 8.7 & 11.4 & 16.1 & 18.7 & 26.5 & 34.9 & 38.4 & 47.2 & 50.3 \\
& & Reward & 0.9 & 1.1 & 1.8 & 2.4 & 3.6 & 4.1 & 5.9 & 7.8 & 8.4 & 10.2 & 10.8 \\
\cmidrule(lr){2-14}
& \multirow{2}{*}{\begin{tabular}[c]{@{}l@{}}Memoryless Policy\\+ MLP Encoder\end{tabular}} & Success (\%) & 1.1 & 2.8 & 6.7 & 8.9 & 13.8 & 18.6 & 24.2 & 31.9 & 35.7 & 45.1 & 47.3 \\
& & Reward & 0.3 & 0.4 & 1.3 & 1.7 & 2.5 & 3.6 & 5.0 & 7.0 & 8.1 & 10.0 & 10.4 \\
\cmidrule(lr){2-14}
& \multirow{2}{*}{Memoryless Policy} & Success (\%) & 1.0 & 2.7 & 6.5 & 8.6 & 13.4 & 18.2 & 23.6 & 31.3 & 35.1 & 44.3 & 46.2 \\
& & Reward & 0.3 & 0.4 & 1.2 & 1.6 & 2.3 & 3.4 & 4.7 & 6.8 & 7.9 & 9.7 & 10.1 \\
\midrule
\multirow{12}{*}{\begin{tabular}[c]{@{}l@{}}Handle-\\Press\end{tabular}} 
& \multirow{2}{*}{\begin{tabular}[c]{@{}l@{}}MEMBOT\\- State Space\end{tabular}} & Success (\%) & \textcolor{blue}{7.0} & \textcolor{blue}{8.0} & \textcolor{blue}{15.0} & \textcolor{blue}{28.0} & \textcolor{blue}{34.0} & \textcolor{blue}{54.0} & \textcolor{blue}{62.0} & \textcolor{blue}{76.0} & \textcolor{blue}{74.0} & \textcolor{blue}{75.0} & \textcolor{blue}{82.0} \\
& & Reward & \textcolor{blue}{1.95} & \textcolor{blue}{1.67} & \textcolor{blue}{4.48} & \textcolor{blue}{7.80} & \textcolor{blue}{9.40} & \textcolor{blue}{15.07} & \textcolor{blue}{18.62} & \textcolor{blue}{24.15} & \textcolor{blue}{22.80} & \textcolor{blue}{24.94} & \textcolor{blue}{28.36} \\
\cmidrule(lr){2-14}
& \multirow{2}{*}{\begin{tabular}[c]{@{}l@{}}MEMBOT\\- LSTM (Full)\end{tabular}} & Success (\%) & 3.2 & 8.4 & 8.3 & 12.7 & 18.5 & 25.2 & 28.1 & 31.8 & 40.2 & 45.3 & 47.9 \\
& & Reward & 1.0 & 2.8 & 3.1 & 4.4 & 6.5 & 8.2 & 9.1 & 10.4 & 12.7 & 14.6 & 14.8 \\
\cmidrule(lr){2-14}
& \multirow{2}{*}{\begin{tabular}[c]{@{}l@{}}MEMBOT\\- LSTM (Partial-10)\end{tabular}} & Success (\%) & 3.1 & 8.5 & 8.4 & 12.3 & 18.3 & 25.0 & 27.9 & 32.7 & 40.0 & 45.1 & 47.8 \\
& & Reward & 1.0 & 2.9 & 3.2 & 4.1 & 6.4 & 8.1 & 8.9 & 10.8 & 12.6 & 14.5 & 14.8 \\
\cmidrule(lr){2-14}
& \multirow{2}{*}{\begin{tabular}[c]{@{}l@{}}MEMBOT\\- LSTM (No Recon)\end{tabular}} & Success (\%) & 2.9 & 7.8 & 7.7 & 12.0 & 17.8 & 24.3 & 27.2 & 32.1 & 39.2 & 44.2 & 47.0 \\
& & Reward & 0.9 & 2.6 & 2.8 & 4.0 & 6.0 & 7.7 & 8.5 & 10.5 & 12.2 & 14.1 & 14.6 \\
\cmidrule(lr){2-14}
& \multirow{2}{*}{\begin{tabular}[c]{@{}l@{}}Memoryless Policy\\+ MLP Encoder\end{tabular}} & Success (\%) & 1.1 & 1.8 & 4.7 & 7.8 & 11.9 & 17.6 & 21.1 & 29.8 & 29.7 & 36.9 & 40.1 \\
& & Reward & 0.1 & 0.5 & 1.2 & 2.0 & 3.1 & 5.0 & 5.7 & 8.9 & 9.0 & 11.9 & 14.1 \\
\cmidrule(lr){2-14}
& \multirow{2}{*}{Memoryless Policy} & Success (\%) & 1.0 & 1.7 & 4.6 & 7.4 & 11.3 & 17.2 & 20.4 & 29.1 & 29.2 & 36.1 & 39.3 \\
& & Reward & 0.1 & 0.5 & 1.1 & 1.9 & 2.9 & 4.7 & 5.4 & 8.6 & 8.7 & 11.6 & 13.8 \\
\midrule
\multirow{12}{*}{\begin{tabular}[c]{@{}l@{}}Drawer-\\Close\end{tabular}} 
& \multirow{2}{*}{\begin{tabular}[c]{@{}l@{}}MEMBOT\\- State Space\end{tabular}} & Success (\%) & 62.0 & 53.0 & 67.0 & 62.0 & 62.0 & 61.0 & 58.0 & 61.0 & 56.0 & 56.0 & 54.0 \\
& & Reward & 15.38 & 13.72 & 19.54 & 17.69 & 18.37 & 18.43 & 18.20 & 19.95 & 18.72 & 19.56 & 18.21 \\
\cmidrule(lr){2-14}
& \multirow{2}{*}{\begin{tabular}[c]{@{}l@{}}MEMBOT\\- LSTM (Full)\end{tabular}} & Success (\%) & \textcolor{blue}{65.8} & \textcolor{blue}{68.7} & \textcolor{blue}{67.3} & \textcolor{blue}{63.9} & \textcolor{blue}{68.6} & \textcolor{blue}{72.1} & \textcolor{blue}{71.5} & \textcolor{blue}{76.3} & \textcolor{blue}{76.9} & \textcolor{blue}{74.3} & \textcolor{blue}{72.1} \\
& & Reward & \textcolor{blue}{23.4} & \textcolor{blue}{24.3} & \textcolor{blue}{24.6} & \textcolor{blue}{22.3} & \textcolor{blue}{24.8} & \textcolor{blue}{26.1} & \textcolor{blue}{26.7} & \textcolor{blue}{27.8} & \textcolor{blue}{28.4} & \textcolor{blue}{28.3} & \textcolor{blue}{26.2} \\
\cmidrule(lr){2-14}
& \multirow{2}{*}{\begin{tabular}[c]{@{}l@{}}MEMBOT\\- LSTM (Partial-10)\end{tabular}} & Success (\%) & 65.1 & 68.9 & 67.0 & 64.2 & 68.4 & 71.8 & 71.6 & 75.8 & 76.5 & 74.0 & 72.0 \\
& & Reward & 23.3 & 24.5 & 24.5 & 22.6 & 24.8 & 26.1 & 26.8 & 27.8 & 28.4 & 28.3 & 26.3 \\
\cmidrule(lr){2-14}
& \multirow{2}{*}{\begin{tabular}[c]{@{}l@{}}MEMBOT\\- LSTM (No Recon)\end{tabular}} & Success (\%) & 64.2 & 67.4 & 66.1 & 63.7 & 67.6 & 70.8 & 70.6 & 74.9 & 75.6 & 72.9 & 71.4 \\
& & Reward & 22.7 & 23.6 & 23.9 & 22.1 & 24.1 & 25.4 & 26.0 & 27.1 & 27.7 & 27.6 & 25.9 \\
\cmidrule(lr){2-14}
& \multirow{2}{*}{\begin{tabular}[c]{@{}l@{}}Memoryless Policy\\+ MLP Encoder\end{tabular}} & Success (\%) & 37.8 & 50.5 & 42.4 & 49.2 & 48.7 & 48.0 & 56.1 & 51.5 & 54.0 & 56.6 & 64.4 \\
& & Reward & 10.0 & 14.4 & 11.6 & 15.0 & 14.9 & 15.1 & 17.8 & 17.5 & 17.6 & 19.4 & 22.9 \\
\cmidrule(lr){2-14}
& \multirow{2}{*}{Memoryless Policy} & Success (\%) & 37.3 & 49.4 & 42.3 & 48.4 & 47.4 & 47.5 & 54.5 & 50.5 & 53.4 & 55.5 & 63.5 \\
& & Reward & 9.8 & 14.1 & 11.5 & 14.8 & 14.5 & 15.0 & 17.3 & 17.2 & 17.4 & 19.0 & 22.6 \\
\bottomrule
\end{tabular}%
}
\end{table}

\subsection{Qualitative Analysis}

As shown in Fig.~\ref{fig:perfvsobs}, we show the evaluation results for three held-out MetaWorld robotic object manipulation tasks: drawer-close, handle-press, and plate-slide. We present the episode success rates and rewards for both memory-based and memoryless agents as a function of observation probability, where lower probabilities indicate more frequent observation dropouts. At full observability (observation probability $p$= 1.0), both memory-based and memoryless agents achieved comparable performance across all tasks, with slightly better performance using memory-based model. For drawer-close (Fig.~\ref{fig:perfvsobs} a1), the performance for MEMBOT and memoryless approaches are very similar. Both approaches reached approximately 50\% success rate with episode rewards around 11. Similar pattern was observed for handle-press task (Fig.~\ref{fig:perfvsobs} b1) and plate-slide (Fig.~\ref{fig:perfvsobs} c1), with success rates of approximately 45\% and 70\% respectively.

We further examine the robustness to the intermittent observations for memory-based model MEMBOT, its variations and memoryless baselines. The memory-based agent MEMBOT with SSM observer demonstrated superior robustness as observation availability decreased. In the drawer-close task (Fig.~\ref{fig:perfvsobs} a1), we can observe that while the memoryless agent's performance degraded substantially below 0.8 observation probability, the memory-based agent MEMBOT maintained relatively stable performance down to 0.6 observation probability. This trend was consistent across handle-press (Fig.~\ref{fig:perfvsobs} b1) and plate-slide (Fig.~\ref{fig:perfvsobs} c1) tasks, though not as evident as the drawer-close task. The relative performance differences (Fig.~\ref{fig:perfvsobs} a2-c2) highlight the memory-based agent's advantage under intermittent POMDP. For the drawer-close task, we can see that the memory-based agent even showed some positive relative improvements in both success rate and reward for some observation probabilities when the observation probability dropped below 0.9. The most pronounced benefits were observed at intermediate observation probabilities (0.6-0.8), where the memory-based agent maintained performance levels closer to full observability conditions while the memoryless agent experienced significant degradation. The magnitude of improvement varied across tasks. The plate-slide task (Fig.~\ref{fig:perfvsobs} c1-c2) showed the most consistent performance for the memory-based agent across all observation probabilities, while drawer-close exhibited the largest performance gaps between approaches at low observation probabilities. The task complexities requires different levels of importance of temporal information for successful completions, which may cause the variations of the performance. 

\subsection{Quantitative Analysis}
To compare the performance of MEMBOT, its variations and memoryless approaches, Table~\ref{tab:comprehensive_ablation} shows comprehensive quantitative results of three held-out tasks. MEMBOT (State-Space) represents the MEMBOT with SSM architecture using the entire temporal history for training; MEMBOT-LSTM (Full) represents the MEMBOT with LSTM architecture using the entire temporal history for training; MEMBOT-LSTM (Partial-10) represents the MEMBOT with LSTM architecture using the latest 10 observations for training, MEMBOT (No Recon) represents MEMBOT architecture without the reconstruction loss; Memoryless Policy + MLP Encoder represents one baseline model using MLP policy with a MLP encoder; Memoryless policy represents one baseline model using MLP policy. 

Across the three held-out tasks, we observed very distinct performance scaling behaviors as observation probability decreased. We can see that the drawer-close task with MEMBOT and its variations maintained relatively high performance even at severe observation dropout (65.8\% success at 0.5 observation probability), while plate-slide and handle-press tasks showed more gradual degradation patterns, dropping to 4.6\% and 3.2\% success rates respectively at the same observation level. We can further define critical performance thresholds to quantify the performance degradation. The critical performance thresholds emerged at different observation probabilities for each task. For handle-press, performance remained below 10\% success until observation probability exceeded 0.65, suggesting a minimum observability requirement. In contrast, drawer-close maintained above 60\% success across all tested observation levels, indicating a larger tolerance to intermittent observations.

MEMBOT demonstrated consistent performance improvements over baselines across the full range of observation probabilities ranging from 0.5 to 1.0. The improvements of performance were most pronounced at lower observation probabilities, with relative gains of up to 220\% in success rate for handle-press at 0.5 observation probability compared to other memoryless baselines. The results reveal a clear task difficulty hierarchy under partial observability conditions. Drawer-close emerged as the most robust task to observation dropout, followed by plate-slide, with handle-press showing the greatest sensitivity to missing observations. This hierarchy remained consistent across all tested methods and observation levels. We can see that metrics success rates and episode rewards showed strong positive correlation across all conditions, all tasks and all methods. This correlation indicate that the sparse reward used in MetaWorld environment effectively captured task completion and that improved success rates translated to proportionally higher cumulative rewards.

The most striking performance difference across tasks that we observe is that different task requires significant different levels of dependencies on temporal historical information. This observation is manifested, for example, by drawer-close task and plate-slide task. The drawer-close task's resilience to observation dropout (maintaining $>$65\% success even at 0.5 observation probability) suggests that drawer manipulation relies heavily on proprioceptive feedback and contact forces that persist between visual observations. In contrast, handle-press and plate-slide tasks showed greater sensitivity to visual intermittency, likely due to their dependence on precise spatial alignment and continuous visual feedback for successful completion. The observation probability thresholds identified in our experiments have practical implications for robotic system design. For handle-press-like tasks requiring $>$65\% observation availability for reasonable performance, system designers must prioritize sensor reliability and redundancy. In contrast, drawer-close-like tasks could operate effectively even with significant sensor failures or bandwidth limitations. The consistent performance scaling patterns suggest that MEMBOT's benefits are predictable and can inform deployment decisions. The approach shows particular promise for environments with intermittent connectivity, sensor occlusions, or computational constraints that necessitate reduced sensing frequency.

\section{Discussion}



We further investigate the architecture of the MEMBOT design and training methods. Based on the Table~\ref{tab:comprehensive_ablation}, our ablation study shows that MEMBOT's superior performance primarily results from its memory-based temporal modeling rather than improved observation encoding alone, since we can see memoryless policy with MLP observation encoder is marginally better than the memoryless policy baseline. The minimal difference between memoryless policies with and without MLP encoders (typically $<$2 percentage points) demonstrates that static feature extraction provides limited benefits under partial observability. In contrast, the memory-based component consistently delivered substantial improvements, particularly at lower observation probabilities, highlighting the critical role of temporal memory in handling intermittent sensing. The reconstruction loss, while providing modest but consistent improvements (0.3--0.5 percentage points), serves as an important regularization mechanism. This auxiliary objective encourages the belief encoder to maintain interpretable latent representations, potentially improving generalization and stability during online fine-tuning. The Partial-10 configuration's comparable performance to the full model suggests that the approach scales efficiently and may not require extensive pretraining data. In MEMBOT implementation, we opt to use LSTM-based architecture. Alternatively, we may employ other memory-based methods with transformer-based architectures. The selection of State-Space Model (SSM) and LSTM emphasize our consideration of lightweight model with the capability of taking streaming inputs while the other transformer-based models requires sequence inputs. In addition, for the sake of the scope projects, we opt to employ the model which does not requires extensive amount of data because we apply multi-task learning and model variations, which needs substantial amount of experiments. 

For ethical concerns, we may expect relatively minimal ethical concerns since this architecture focuses on improving the robustness of the planning and control system of robots in presence of intermittent POMDPs which aims at addressing sensory system malfunctions or delays. However, we still need attention to extrapolation of temporal latent information which may not be representative of the current state. Leveraging these condensed hidden information may be beneficial to robot agents and may also cause unreliable control signals. In addition, increased autonomy also raises concerns about system transparency, accountability, and the potential for misuse in military or surveillance applications. The ability to operate effectively with limited sensory information could potentially enable autonomous systems to function in scenarios where human oversight is reduced, raising questions about appropriate safeguards and human-in-the-loop mechanisms. Future development should carefully consider these ethical implications and incorporate appropriate safety measures.

For future work, we've identified some potential limitations of our current approaches. For example, fixed LSTM size may limit scalability to long-horizon tasks. Current models tested only on low-dimensional, selected tasks. In future work, we will replacing SSM/LSTM with Mamba to improve sequence modeling~\citep{gu2023mamba}. Training LSTM/SSM with different observation probability to improve generalization with further experimentation with architecture size, evaluation tests, ablation studies, sensitivity analysis and hyperparameter tuning. We will further extend to visual inputs and evaluate on real robots with intermittent sensing. In addition, we will further extend the training environmental steps. Given the substantial amount of training requirement, we have only run 100,000 environment steps which may lead to a suboptimal policy.

\section{Conclusion}
We develop MEMBOT, a memory-based approach for robotic object manipulation, to address intermittent observations due to  due to hardware malfunctions, environment realities, physical obstruction or system optimization trade-offs. MEMBOT primarily consists of three modules, including a shared task-agnostic observer module, a task-specific actor and a task-specific ensemble of critics. The MEMBOT training process consists of two phases, including an offline pretraining phase and an online finetuning phase. During offline pretraining, we aim to pretrain a task-agnostic observer to produce latent belief states, leveraging the hidden states of SSM and hidden states and cell states of LSTM. During the online finetuning state, we aim to finetune the observer and train the task-specific actor and critics. The shared observer is designed to improve the online training for policy optimization. 

Through comprehensive experimental evaluation across multiple manipulation tasks and architecture variations, we demonstrate that MEMBOT significantly outperforms existing baseline methods, maintaining robust performance even under severe observation dropout conditions up to the observation probability of 0.5. The key contributions of this work include: (1) a modular architecture that separates belief state learning from policy optimization, enabling stable training and effective transfer learning; (2) a two-phase training methodology that combines behavior cloning with reconstruction objectives to learn robust latent temporal representations; (3) empirical validation showing 2-3x improvements in sample efficiency for transfer learning and maintenance of 65-80\% performance under 50\% observation availability; and (4) systematic ablation studies identifying the critical components for achieving robust performance.

\section{Team Contributions}
\begin{itemize}
  \item \textbf{Eyan Noronha:} Memoryless baseline models (x2), Memeory-based models using Mamba (CPU) and State-Space Model (SSM), Offline training and online training (x5 Metaworld tasks), literature review, proposal/milestone/final report. 
  \item \textbf{Yousef Liang:} Memory-based models using LSTM and variations (x3), Offline multi-task learning (x5 Robomic task) for pretraining and online transfer learning (x3 Metaworld tasks), literature review, proposal/milestone/final report. 
\end{itemize}

\paragraph{Changes from Proposal} In our original proposal, we use sequence models to develop memory-based policies using vision-language models (VLMs) as reward functions for robotic reinforcement learning. In our further design and implementation, we use sequence models to develop memory-based policies using predefined tasks and reward functions for robotic reinforcement learning which leverages the MetaWorld environment and Robomimic datasets. We've made this change from our proposal, aiming to decouple the effects of using VLM-based rewards from those of developing memory-based policies. Our immediate focus is on building robust memory-based policies with predefined reward functions.

\bibliographystyle{ACM-Reference-Format}
\bibliography{main}


\begin{thebibliography}{18}


\ifx \showCODEN    \undefined \def \showCODEN     #1{\unskip}     \fi
\ifx \showISBNx    \undefined \def \showISBNx     #1{\unskip}     \fi
\ifx \showISBNxiii \undefined \def \showISBNxiii  #1{\unskip}     \fi
\ifx \showISSN     \undefined \def \showISSN      #1{\unskip}     \fi
\ifx \showLCCN     \undefined \def \showLCCN      #1{\unskip}     \fi
\ifx \shownote     \undefined \def \shownote      #1{#1}          \fi
\ifx \showarticletitle \undefined \def \showarticletitle #1{#1}   \fi
\ifx \showURL      \undefined \def \showURL       {\relax}        \fi
\providecommand\bibfield[2]{#2}
\providecommand\bibinfo[2]{#2}
\providecommand\natexlab[1]{#1}
\providecommand\showeprint[2][]{arXiv:#2}

\bibitem[Aikins et~al\mbox{.}(2024)]%
        {aikins2024robust}
\bibfield{author}{\bibinfo{person}{Godwyll Aikins}, \bibinfo{person}{Sagar Jagtap}, {and} \bibinfo{person}{Kim-Doang Nguyen}.} \bibinfo{year}{2024}\natexlab{}.
\newblock \showarticletitle{A robust strategy for uav autonomous landing on a moving platform under partial observability}.
\newblock \bibinfo{journal}{\emph{Drones}} \bibinfo{volume}{8}, \bibinfo{number}{6} (\bibinfo{year}{2024}), \bibinfo{pages}{232}.
\newblock


\bibitem[Avrachenkov et~al\mbox{.}(2025)]%
        {avrachenkov2025constrained}
\bibfield{author}{\bibinfo{person}{Konstantin Avrachenkov}, \bibinfo{person}{Madhu Dhiman}, {and} \bibinfo{person}{Veeraruna Kavitha}.} \bibinfo{year}{2025}\natexlab{}.
\newblock \showarticletitle{Constrained Average-Reward Intermittently Observable MDPs}.
\newblock \bibinfo{journal}{\emph{arXiv preprint arXiv:2504.13823}} (\bibinfo{year}{2025}).
\newblock


\bibitem[Basich et~al\mbox{.}(2022)]%
        {basich2022planning}
\bibfield{author}{\bibinfo{person}{Connor Basich}, \bibinfo{person}{John Peterson}, {and} \bibinfo{person}{Shlomo Zilberstein}.} \bibinfo{year}{2022}\natexlab{}.
\newblock \showarticletitle{Planning with intermittent state observability: Knowing when to act blind}. In \bibinfo{booktitle}{\emph{2022 IEEE/RSJ International Conference on Intelligent Robots and Systems (IROS)}}. IEEE, \bibinfo{pages}{11657--11664}.
\newblock


\bibitem[Becker et~al\mbox{.}(2024)]%
        {becker2024kalmamba}
\bibfield{author}{\bibinfo{person}{Philipp Becker}, \bibinfo{person}{Niklas Freymuth}, {and} \bibinfo{person}{Gerhard Neumann}.} \bibinfo{year}{2024}\natexlab{}.
\newblock \showarticletitle{KalMamba: Towards Efficient Probabilistic State Space Models for RL under Uncertainty}.
\newblock \bibinfo{journal}{\emph{arXiv preprint arXiv:2406.15131}} (\bibinfo{year}{2024}).
\newblock


\bibitem[Cho et~al\mbox{.}(2023)]%
        {cho2023low}
\bibfield{author}{\bibinfo{person}{Jae-Kyung Cho}, \bibinfo{person}{Chan Kim}, \bibinfo{person}{Mohamed Khalid~M Jaffar}, \bibinfo{person}{Michael~W Otte}, {and} \bibinfo{person}{Seong-Woo Kim}.} \bibinfo{year}{2023}\natexlab{}.
\newblock \showarticletitle{Low-level controller in response to changes in quadrotor dynamics}. In \bibinfo{booktitle}{\emph{2023 IEEE International Conference on Robotics and Automation (ICRA)}}. IEEE, \bibinfo{pages}{5317--5323}.
\newblock


\bibitem[Fang et~al\mbox{.}(2019)]%
        {fang2019scene}
\bibfield{author}{\bibinfo{person}{Kuan Fang}, \bibinfo{person}{Alexander Toshev}, \bibinfo{person}{Li Fei-Fei}, {and} \bibinfo{person}{Silvio Savarese}.} \bibinfo{year}{2019}\natexlab{}.
\newblock \showarticletitle{Scene memory transformer for embodied agents in long-horizon tasks}. In \bibinfo{booktitle}{\emph{Proceedings of the IEEE/CVF conference on computer vision and pattern recognition}}. \bibinfo{pages}{538--547}.
\newblock


\bibitem[Gu and Dao(2023)]%
        {gu2023mamba}
\bibfield{author}{\bibinfo{person}{Albert Gu} {and} \bibinfo{person}{Tri Dao}.} \bibinfo{year}{2023}\natexlab{}.
\newblock \showarticletitle{Mamba: Linear-time sequence modeling with selective state spaces}.
\newblock \bibinfo{journal}{\emph{arXiv preprint arXiv:2312.00752}} (\bibinfo{year}{2023}).
\newblock


\bibitem[Haarnoja et~al\mbox{.}(2018)]%
        {haarnoja2018soft}
\bibfield{author}{\bibinfo{person}{Tuomas Haarnoja}, \bibinfo{person}{Aurick Zhou}, \bibinfo{person}{Pieter Abbeel}, {and} \bibinfo{person}{Sergey Levine}.} \bibinfo{year}{2018}\natexlab{}.
\newblock \showarticletitle{Soft actor-critic: Off-policy maximum entropy deep reinforcement learning with a stochastic actor}. In \bibinfo{booktitle}{\emph{International conference on machine learning}}. Pmlr, \bibinfo{pages}{1861--1870}.
\newblock


\bibitem[Han(2023)]%
        {han2023multimodal}
\bibfield{author}{\bibinfo{person}{Zhuqin Han}.} \bibinfo{year}{2023}\natexlab{}.
\newblock \showarticletitle{Multimodal intelligent logistics robot combining 3D CNN, LSTM, and visual SLAM for path planning and control}.
\newblock \bibinfo{journal}{\emph{Frontiers in Neurorobotics}}  \bibinfo{volume}{17} (\bibinfo{year}{2023}), \bibinfo{pages}{1285673}.
\newblock


\bibitem[Hausknecht and Stone(2015)]%
        {hausknecht2015deep}
\bibfield{author}{\bibinfo{person}{Matthew~J Hausknecht} {and} \bibinfo{person}{Peter Stone}.} \bibinfo{year}{2015}\natexlab{}.
\newblock \showarticletitle{Deep Recurrent Q-Learning for Partially Observable MDPs.}. In \bibinfo{booktitle}{\emph{AAAI fall symposia}}, Vol.~\bibinfo{volume}{45}. \bibinfo{pages}{141}.
\newblock


\bibitem[Jiang et~al\mbox{.}(2022)]%
        {jiang2022multi}
\bibfield{author}{\bibinfo{person}{Peng Jiang}, \bibinfo{person}{Jiafeng Ma}, \bibinfo{person}{Zhiwen Zhang}, {and} \bibinfo{person}{Jianming Zhang}.} \bibinfo{year}{2022}\natexlab{}.
\newblock \showarticletitle{Multi-Sensor Fusion Framework for Obstacle Avoidance via Deep Reinforcement Learning}. In \bibinfo{booktitle}{\emph{2022 2nd International Conference on Electrical Engineering and Control Science (IC2ECS)}}. IEEE, \bibinfo{pages}{153--156}.
\newblock


\bibitem[K{\"a}stner et~al\mbox{.}(2021)]%
        {kastner2021autonomous}
\bibfield{author}{\bibinfo{person}{Linh K{\"a}stner}, \bibinfo{person}{Zhengcheng Shen}, \bibinfo{person}{Cornelius Marx}, {and} \bibinfo{person}{Jens Lambrecht}.} \bibinfo{year}{2021}\natexlab{}.
\newblock \showarticletitle{Autonomous navigation in complex environments using memory-aided deep reinforcement learning}. In \bibinfo{booktitle}{\emph{2021 IEEE/SICE International Symposium on System Integration (SII)}}. IEEE, \bibinfo{pages}{170--175}.
\newblock


\bibitem[Li et~al\mbox{.}(2024)]%
        {li2024deep}
\bibfield{author}{\bibinfo{person}{Yufeng Li}, \bibinfo{person}{Jian Gao}, \bibinfo{person}{Yimin Chen}, \bibinfo{person}{Yaozhen He}, {and} \bibinfo{person}{Boxu Min}.} \bibinfo{year}{2024}\natexlab{}.
\newblock \showarticletitle{Deep Reinforcement Learning-Based End-to-End Navigation of Mobile Robots with Reward Shaping}. In \bibinfo{booktitle}{\emph{2024 IEEE 22nd International Conference on Industrial Informatics (INDIN)}}. IEEE, \bibinfo{pages}{1--6}.
\newblock


\bibitem[Mandlekar et~al\mbox{.}(2021)]%
        {mandlekar2021matters}
\bibfield{author}{\bibinfo{person}{Ajay Mandlekar}, \bibinfo{person}{Danfei Xu}, \bibinfo{person}{Josiah Wong}, \bibinfo{person}{Soroush Nasiriany}, \bibinfo{person}{Chen Wang}, \bibinfo{person}{Rohun Kulkarni}, \bibinfo{person}{Li Fei-Fei}, \bibinfo{person}{Silvio Savarese}, \bibinfo{person}{Yuke Zhu}, {and} \bibinfo{person}{Roberto Mart{\'\i}n-Mart{\'\i}n}.} \bibinfo{year}{2021}\natexlab{}.
\newblock \showarticletitle{What matters in learning from offline human demonstrations for robot manipulation}.
\newblock \bibinfo{journal}{\emph{arXiv preprint arXiv:2108.03298}} (\bibinfo{year}{2021}).
\newblock


\bibitem[Parsbin and Akraminia(2023)]%
        {parsbin2023visual}
\bibfield{author}{\bibinfo{person}{Ali Parsbin} {and} \bibinfo{person}{Mahdi Akraminia}.} \bibinfo{year}{2023}\natexlab{}.
\newblock \showarticletitle{Visual Navigation for Obstacle Avoidance Using Deep Reinforcement Learning with Policy Optimization}. In \bibinfo{booktitle}{\emph{2023 11th RSI International Conference on Robotics and Mechatronics (ICRoM)}}. IEEE, \bibinfo{pages}{153--158}.
\newblock


\bibitem[Yu et~al\mbox{.}(2020)]%
        {yu2020meta}
\bibfield{author}{\bibinfo{person}{Tianhe Yu}, \bibinfo{person}{Deirdre Quillen}, \bibinfo{person}{Zhanpeng He}, \bibinfo{person}{Ryan Julian}, \bibinfo{person}{Karol Hausman}, \bibinfo{person}{Chelsea Finn}, {and} \bibinfo{person}{Sergey Levine}.} \bibinfo{year}{2020}\natexlab{}.
\newblock \showarticletitle{Meta-world: A benchmark and evaluation for multi-task and meta reinforcement learning}. In \bibinfo{booktitle}{\emph{Conference on robot learning}}. PMLR, \bibinfo{pages}{1094--1100}.
\newblock


\bibitem[Zhang et~al\mbox{.}(2015)]%
        {zhang2015policy}
\bibfield{author}{\bibinfo{person}{Marvin Zhang}, \bibinfo{person}{Sergey Levine}, \bibinfo{person}{Zoe McCarthy}, \bibinfo{person}{Chelsea Finn}, {and} \bibinfo{person}{Pieter Abbeel}.} \bibinfo{year}{2015}\natexlab{}.
\newblock \showarticletitle{Policy learning with continuous memory states for partially observed robotic control}.
\newblock \bibinfo{journal}{\emph{CoRR, abs/1507.01273}} (\bibinfo{year}{2015}).
\newblock


\bibitem[Zhong et~al\mbox{.}(2023)]%
        {zhong2023optimizing}
\bibfield{author}{\bibinfo{person}{Patrick Zhong}, \bibinfo{person}{Federico Rossi}, {and} \bibinfo{person}{Dylan~A Shell}.} \bibinfo{year}{2023}\natexlab{}.
\newblock \showarticletitle{Optimizing pre-scheduled, intermittently-observed MDPs}. In \bibinfo{booktitle}{\emph{2023 62nd IEEE Conference on Decision and Control (CDC)}}. IEEE, \bibinfo{pages}{2244--2251}.
\newblock


\end{thebibliography}

\newpage

\appendix

\section{Additional Results}
\begin{table}[h!]
\centering
\caption{Relative performance degradation compared to full observability (obs prob = 1.0). The best results are colored in blue. }
\label{tab:relative_degradation}
\resizebox{\textwidth}{!}{%
\begin{tabular}{llc|ccccccccccc}
\toprule
\multirow{2}{*}{Task} & \multirow{2}{*}{Method} & \multirow{2}{*}{Metric} & \multicolumn{10}{c}{Observation Probability} \\
\cmidrule(lr){4-13}
& & & 0.5 & 0.55 & 0.60 & 0.65 & 0.70 & 0.75 & 0.80 & 0.85 & 0.90 & 0.95 \\
\midrule
\multirow{12}{*}{\begin{tabular}[c]{@{}l@{}}Plate-\\Slide\end{tabular}} 
& \multirow{2}{*}{\begin{tabular}[c]{@{}l@{}}MEMBOT\\- State Space\end{tabular}} & Success (\%) & \textcolor{blue}{-90.5} & -92.6 & -90.5 & -77.9 & -74.7 & \textcolor{blue}{-56.8} & -52.6 & \textcolor{blue}{-26.3} & \textcolor{blue}{-20.0} & -11.6 \\
& & Reward (\%) & -94.3 & -95.6 & -92.0 & -85.6 & -79.9 & -68.5 & -60.8 & -37.0 & -28.9 & -18.6 \\
\cmidrule(lr){2-13}
& \multirow{2}{*}{\begin{tabular}[c]{@{}l@{}}MEMBOT\\- LSTM (Full)\end{tabular}} & Success (\%) & -91.0 & -91.6 & \textcolor{blue}{-82.0} & \textcolor{blue}{-76.8} & \textcolor{blue}{-67.4} & -64.1 & \textcolor{blue}{-46.7} & -30.1 & -23.6 & -5.7 \\
& & Reward (\%) & -91.0 & -91.0 & \textcolor{blue}{-82.0} & \textcolor{blue}{-76.6} & \textcolor{blue}{-64.9} & -63.1 & \textcolor{blue}{-43.2} & \textcolor{blue}{-27.0} & -21.6 & -5.4 \\
\cmidrule(lr){2-13}
& \multirow{2}{*}{\begin{tabular}[c]{@{}l@{}}MEMBOT\\- LSTM (Partial-10)\end{tabular}} & Success (\%) & -90.8 & -91.8 & -82.2 & -76.9 & -67.5 & -62.2 & \textcolor{blue}{-46.7} & -30.0 & -23.1 & -5.7 \\
& & Reward (\%) & \textcolor{blue}{-90.9} & -90.9 & -82.7 & -77.3 & -65.5 & \textcolor{blue}{-60.0} & -43.6 & -27.3 & \textcolor{blue}{-20.0} & -5.5 \\
\cmidrule(lr){2-13}
& \multirow{2}{*}{\begin{tabular}[c]{@{}l@{}}MEMBOT\\- LSTM (No Recon)\end{tabular}} & Success (\%) & -91.6 & \textcolor{blue}{-90.5} & -82.7 & -77.3 & -68.0 & -62.8 & -47.3 & -30.6 & -23.7 & -6.2 \\
& & Reward (\%) & -91.7 & \textcolor{blue}{-89.8} & -83.3 & -77.8 & -66.7 & -62.0 & -45.4 & -27.8 & -22.2 & -5.6 \\
\cmidrule(lr){2-13}
& \multirow{2}{*}{\begin{tabular}[c]{@{}l@{}}Memoryless Policy\\+ MLP Encoder\end{tabular}} & Success (\%) & -97.7 & -94.1 & -85.8 & -81.2 & -70.8 & -60.7 & -48.8 & -32.6 & -24.5 & -4.6 \\
& & Reward (\%) & -97.1 & -96.2 & -87.5 & -83.7 & -76.0 & -65.4 & -51.9 & -32.7 & -22.1 & \textcolor{blue}{-3.8} \\
\cmidrule(lr){2-13}
& \multirow{2}{*}{Memoryless Policy} & Success (\%) & -97.8 & -94.2 & -85.9 & -81.4 & -71.0 & -60.6 & -48.9 & -32.3 & -24.0 & \textcolor{blue}{-4.1} \\
& & Reward (\%) & -97.0 & -96.0 & -88.1 & -84.2 & -77.2 & -66.3 & -53.5 & -32.7 & -21.8 & -4.0 \\
\midrule
\multirow{12}{*}{\begin{tabular}[c]{@{}l@{}}Handle-\\Press\end{tabular}} 
& \multirow{2}{*}{\begin{tabular}[c]{@{}l@{}}MEMBOT\\- State Space\end{tabular}} & Success (\%) & \textcolor{blue}{-91.5} & -90.2 & \textcolor{blue}{-81.7} & \textcolor{blue}{-65.9} & \textcolor{blue}{-58.5} & \textcolor{blue}{-34.1} & \textcolor{blue}{-24.4} & \textcolor{blue}{-7.3} & \textcolor{blue}{-9.8} & -8.5 \\
& & Reward (\%) & \textcolor{blue}{-93.1} & -94.1 & -84.2 & -72.5 & -66.9 & -46.9 & \textcolor{blue}{-34.3} & \textcolor{blue}{-14.9} & -19.6 & -12.1 \\
\cmidrule(lr){2-13}
& \multirow{2}{*}{\begin{tabular}[c]{@{}l@{}}MEMBOT\\- LSTM (Full)\end{tabular}} & Success (\%) & -93.3 & -82.5 & -82.7 & -73.5 & -61.4 & -47.4 & -41.3 & -33.6 & -16.1 & \textcolor{blue}{-5.4} \\
& & Reward (\%) & -93.2 & -81.1 & -79.1 & \textcolor{blue}{-70.3} & \textcolor{blue}{-56.1} & \textcolor{blue}{-44.6} & -38.5 & -29.7 & \textcolor{blue}{-14.2} & \textcolor{blue}{-1.4} \\
\cmidrule(lr){2-13}
& \multirow{2}{*}{\begin{tabular}[c]{@{}l@{}}MEMBOT\\- LSTM (Partial-10)\end{tabular}} & Success (\%) & -93.5 & \textcolor{blue}{-82.2} & -82.4 & -74.3 & -61.7 & -47.7 & -41.6 & -31.5 & -16.3 & -5.6 \\
& & Reward (\%) & -93.2 & \textcolor{blue}{-80.4} & \textcolor{blue}{-78.4} & -72.3 & -56.8 & -45.3 & -39.9 & -27.0 & -14.9 & -2.0 \\
\cmidrule(lr){2-13}
& \multirow{2}{*}{\begin{tabular}[c]{@{}l@{}}MEMBOT\\- LSTM (No Recon)\end{tabular}} & Success (\%) & -93.8 & -83.4 & -83.6 & -74.5 & -62.1 & -48.3 & -42.1 & -31.7 & -16.6 & -6.0 \\
& & Reward (\%) & -93.8 & -82.2 & -80.8 & -72.6 & -58.9 & -47.3 & -41.8 & -28.1 & -16.4 & -3.4 \\
\cmidrule(lr){2-13}
& \multirow{2}{*}{\begin{tabular}[c]{@{}l@{}}Memoryless Policy\\+ MLP Encoder\end{tabular}} & Success (\%) & -97.3 & -95.5 & -88.3 & -80.5 & -70.3 & -56.1 & -47.4 & -25.7 & -25.9 & -8.0 \\
& & Reward (\%) & -99.3 & -96.5 & -91.5 & -85.8 & -78.0 & -64.5 & -59.6 & -36.9 & -36.2 & -15.6 \\
\cmidrule(lr){2-13}
& \multirow{2}{*}{Memoryless Policy} & Success (\%) & -97.5 & -95.7 & -88.3 & -81.2 & -71.2 & -56.2 & -48.1 & -25.9 & -25.7 & -8.1 \\
& & Reward (\%) & -99.3 & -96.4 & -92.0 & -86.2 & -79.0 & -65.9 & -60.9 & -37.7 & -37.0 & -15.9 \\
\midrule
\multirow{12}{*}{\begin{tabular}[c]{@{}l@{}}Drawer-\\Close\end{tabular}} 
& \multirow{2}{*}{\begin{tabular}[c]{@{}l@{}}MEMBOT\\- State Space\end{tabular}} & Success (\%) & \textcolor{blue}{14.8} & \textcolor{blue}{-1.9} & \textcolor{blue}{24.1} & \textcolor{blue}{14.8} & \textcolor{blue}{14.8} & \textcolor{blue}{13.0} & \textcolor{blue}{7.4} & \textcolor{blue}{13.0} & 3.7 & \textcolor{blue}{3.7} \\
& & Reward (\%) & -15.5 & -24.7 & \textcolor{blue}{7.3} & \textcolor{blue}{-2.9} & \textcolor{blue}{0.9} & \textcolor{blue}{1.2} & 0.0 & \textcolor{blue}{9.6} & 2.8 & 7.4 \\
\cmidrule(lr){2-13}
& \multirow{2}{*}{\begin{tabular}[c]{@{}l@{}}MEMBOT\\- LSTM (Full)\end{tabular}} & Success (\%) & -8.7 & -4.7 & -6.7 & -11.4 & -4.9 & 0.0 & -0.8 & 5.8 & \textcolor{blue}{6.7} & 3.1 \\
& & Reward (\%) & \textcolor{blue}{-10.7} & -7.3 & -6.1 & -14.9 & -5.3 & -0.4 & \textcolor{blue}{1.9} & 6.1 & \textcolor{blue}{8.4} & \textcolor{blue}{8.0} \\
\cmidrule(lr){2-13}
& \multirow{2}{*}{\begin{tabular}[c]{@{}l@{}}MEMBOT\\- LSTM (Partial-10)\end{tabular}} & Success (\%) & -9.6 & -4.3 & -6.9 & -10.8 & -5.0 & -0.3 & -0.6 & 5.3 & 6.3 & 2.8 \\
& & Reward (\%) & -11.4 & \textcolor{blue}{-6.8} & -6.8 & -14.1 & -5.7 & -0.8 & \textcolor{blue}{1.9} & 5.7 & 8.0 & 7.6 \\
\cmidrule(lr){2-13}
& \multirow{2}{*}{\begin{tabular}[c]{@{}l@{}}MEMBOT\\- LSTM (No Recon)\end{tabular}} & Success (\%) & -10.1 & -5.6 & -7.4 & -10.8 & -5.3 & -0.8 & -1.1 & 4.9 & 5.9 & 2.1 \\
& & Reward (\%) & -12.4 & -8.9 & -7.7 & -14.7 & -7.0 & -1.9 & 0.4 & 4.6 & 6.9 & 6.6 \\
\cmidrule(lr){2-13}
& \multirow{2}{*}{\begin{tabular}[c]{@{}l@{}}Memoryless Policy\\+ MLP Encoder\end{tabular}} & Success (\%) & -41.3 & -21.6 & -34.2 & -23.6 & -24.4 & -25.5 & -12.9 & -20.1 & -16.1 & -12.1 \\
& & Reward (\%) & -56.3 & -37.1 & -49.3 & -34.5 & -34.9 & -34.1 & -22.3 & -23.6 & -23.1 & -15.3 \\
\cmidrule(lr){2-13}
& \multirow{2}{*}{Memoryless Policy} & Success (\%) & -41.3 & -22.2 & -33.4 & -23.8 & -25.4 & -25.2 & -14.2 & -20.5 & -15.9 & -12.6 \\
& & Reward (\%) & -56.6 & -37.6 & -49.1 & -34.5 & -35.8 & -33.6 & -23.5 & -23.9 & -23.0 & -15.9 \\
\bottomrule
\end{tabular}%
}
\end{table}


\end{document}